\documentclass{article}

\PassOptionsToPackage{numbers, compress}{natbib}

\usepackage[arxiv]{nips_2016}
\usepackage{booktabs}
\usepackage{framed} 
\usepackage{amsmath}
\usepackage{algorithm}
\usepackage{algorithmic}
\usepackage{graphicx}
\usepackage{tikz}
\usetikzlibrary{fit}
\usepackage{color, colortbl}
\definecolor{LightCyan}{rgb}{0.88,1,1}
\usepackage[utf8]{inputenc} 
\usepackage[T1]{fontenc}    
\usepackage{hyperref}       
\usepackage{url}            
\usepackage{amsfonts}       
\usepackage{nicefrac}       
\usepackage{microtype}      

\title{Dynamic Collaborative Filtering with Compound Poisson Factorization}

\author{
Ghassen Jerfel \\
Department of Computer Science\\
Princeton University\\
Princeton, NJ 08544 \\
\texttt{gjerfel@princeton.edu} \\
\And
Mehmet E. Basbug \\
Department of Computer Science\\
Princeton University\\
Princeton, NJ 08544 \\
\texttt{mehmetbasbug@yahoo.com} \\
\AND
Barbara E Engelhardt \\
Department of Computer Science\\
Center for Statistics and Machine Learning\\
Princeton University\\
Princeton, NJ 08540 \\
\texttt{bee@princeton.edu} \\
}

\begin{document}

\maketitle
\begin{abstract}
Model-based collaborative filtering analyzes user-item interactions to infer latent factors that represent user preferences and item characteristics in order to predict future interactions. Most collaborative filtering algorithms assume that these latent factors are static, although it has been shown that user preferences and item perceptions drift over time. In this paper, we propose a conjugate and numerically stable dynamic matrix factorization (DCPF) based on compound Poisson matrix factorization that models the smoothly drifting latent factors using Gamma-Markov chains. We propose a numerically stable Gamma chain construction, and then present a stochastic variational inference approach to estimate the parameters of our model. We apply our model to time-stamped ratings data sets: Netflix, Yelp, and Last.fm, where DCPF achieves a higher predictive accuracy than state-of-the-art static and dynamic factorization models.
\end{abstract}	
\section{Introduction}
Collaborative filtering (CF) \citep{handbook} is a popular framework for recommender systems to personalize item recommendations for each customer based on their previous item interactions; these approaches are used by large online platforms where user decision-making is influenced by item recommendations, including Amazon \citep{amazon}, Spotify, and Netflix \citep{koren2009bellkor}. Latent factor models have recently become a research focus within CF. These models represent user preferences and item attributes in terms of latent factors that exploit the low-dimensional latent structure to impute missing user--item interactions~\citep{handbook}. In this context, Poisson factorization models (PF)~\citep{cemgil,hpf,dpf,hcpf,prem2} have demonstrated better accuracy in predicting held-out user--item ratings than other CF models such as Probabilistic Matrix Factorization (PMF) \citep{pmf}, a Gaussian factorization model.

Poisson factorization models for CF generally assume that user preferences and item attributes are static over time \citep{cemgil}. However, in real-world scenarios, user interests and item attributes drift in time \citep{dpf,dynamicGausian}. Items' popularity and reception continuously evolve as new items and categories emerge and global environments change. Customers' preferences and needs change over time, leading them to interact with items in a time-varying way. 
However, static PF models may mistakenly recommend sports documentaries to a current fan of horror movies, or suggest rock songs to a new fan of electronic music. Similarly, they might not recognize that a 50's statistical physics paper~\citep{robbins} is again relevant but to a new machine learning audience. Thus, the static assumption of traditional PF models is not well-aligned with the true generative model of ratings in real-world scenarios.

Temporal dynamics have been explored in the context of other collaborative filtering frameworks. Early approaches underweighted older user--item interactions \citep{weight}. TimeSVD++ proposed an extension to singular value decomposition (SVD) by augmenting matrix factorization with a user--item--time bias term and user-evolving factors, but ignored the drifting nature of the item attributes \citep{timesvd}. Nevertheless, TimeSVD++ was one of the winning algorithms of the Netflix Prize in 2009, motivating the need for further work on temporal dynamics in CF models. Bayesian Poisson tensor factorization proposes an independent latent factor for time (or context) then factorizes a user--item--time tensor \citep{tensor}. This approach models the user and item factors at each time index independently from previous ones, which makes evaluating trends of specific user and item factors impossible. 

Several dynamic extensions to PMF have also been proposed by leveraging a Gaussian state space model to represent the evolving user and item latent factors \citep{gaussian,StateSpace,dynamicGausian}. However, the conjugate gamma-Poisson structure of Poisson factorization models ensures computationally tractable approaches for high-dimensional data using variational inference~\citep{cemgil}. Additionally, long tailed gamma priors powerfully capture the overdispersed user and item behavior in sparse matrices, which leads to better predictive results---in terms of precision and recall---than PMF \citep{pmf} and latent Dirichlet allocation \citep{lda}, among other CF approaches as shown by \citet{hpf}.  

In this paper, we address the problem of evolving user and item factors within the context of Poisson factorization. We propose a novel dynamic PF model: dynamic compound-Poisson factorization (DCPF). DCPF is a novel dynamic probabilistic model that represents the user and item latent factors as independent smoothly-evolving gamma-Markov chains. There has been a recent dynamic extension attempt for PF replacing the gamma priors with a Gaussian state space model ~\citep{dpf}. 
However, this approach compromises the conjugacy of the model, leading to computational intractability or, alternatively, inaccurate numerical approximations for posterior inference. In fact, the lack of conjugacy prevents simple closed-form updates and consequently prevents convenient extensions to the model. Furthermore, Gaussian priors at each time step fail to capture the empirical response distribution and the long-tailed Gamma distributions, as demonstrated by \citet{hpf}.

However, conjugacy cannot be attained with simple gamma chains in a Poisson factorization model. Therefore, we introduce auxiliary gamma chains to each chain of user and item factors to guarantee the conjugacy of the user and item chains within the model. We also introduce a compound Poisson generating distribution \citep{hcpf}. The Poisson compounding allows us to rescale the non-negative multiplicative gamma chains, which would otherwise grow uncontrollably and thus lead to numerical instability during inference. It also provides more flexibility with ratings data types.

In Section \ref{gamma}, we propose a conjugate gamma chain construction with dampened, positively correlated states (i.e., slowly and smoothly evolving in time). In Section \ref{model}, we describe the generative model for DCPF and the stochastic variational inference approach that allows the inference of DCPF to scale to large, sparse data sets. In Section \ref{results}, we present the experimental setup including the data sets, the static and dynamic baselines, the comparison metrics, our hyperparameter initialization choices, and discuss our results. We conclude with potential improvements.
\section{Conjugate and Stable Gamma Markov Chains}
\label{gamma}
In our approach to incorporating smooth temporal dynamics in Poisson factorization, we preserved the gamma-Poisson conjugate structure by placing gamma-Markov chains as priors on the latent factors associated with each user and item. Conjugacy guarantees closed form updates to infer the model in a scalable fashion, and combine it with other conjugate models in the future. Additionally, long tail gamma priors are able to capture diverse sets of users and items in terms of their activity and popularity \citep{hpf}. This allows us to better represent data where a few users might be proactive reviewers while the majority rarely contribute, for example.

We assume that the latent factors evolve smoothly in time and remain non-negative, which requires the gamma chain states to be positively correlated (i.e. each state matches the previous state's trajectory with a higher probability). Therefore, we constructed a Gamma chain by conditioning the state at every time step on the inverse of the state at the previous time step via the gamma scale parameters. This is different from the \citet{acharya} approach of chaining on the shape parameter with no consideration to the temporal smoothness as they used the chains for the substantially different problem of factor analysis.

A simple and intuitive construction of this chain would draw the $k$th component of the user $m$'s latent factor at time $t$ as follows : $u_{m,k,t} \sim Ga(a, b\cdot u_{m,k,t-1})$.

Here, $Ga$ is the gamma distribution with a fixed shape parameter $a$ and a rate parameter $b\cdot u_{m,k,t-1}$.

However, this construction is non-conjugate because the full conditional distribution of the latent factor value, $p(u_{m,k,t}|u_{m,k,t-1},u_{m,k,t+1})$, includes the terms $u$, $1/u$ and $log\;u$. A conjugate density should be in the same distribution family as $p(u_{m,k,t})$ (gamma). However, one can easily check that a gamma density for $u$ cannot be expressed with the above terms.

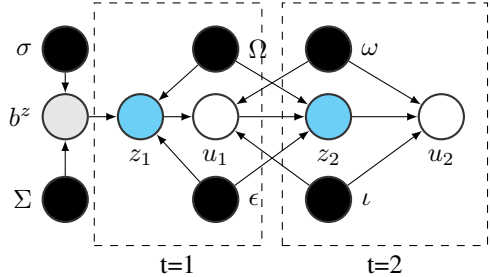
\begin{figure}[!hb]
	\centering
	\begin{tikzpicture}
	\tikzstyle{main}=[circle, minimum size = 6mm, thick, draw =black!80, node distance = 10mm]
	\tikzstyle{connect}=[-latex]
	\tikzstyle{box}=[rectangle, draw=black!100]
	\node[main, fill=cyan!50] (Z) [label=below:$z_1$] at (3,1.1) { };
	\node[main, fill=white] (Ut) [label=below:$u_1$] at(4,1.1) { };
	\node[main, fill=white] (U_{t+1}) [label=below:$u_2$] at(7,1.1) { };
	\node[main, fill=cyan!50] (Zt) [label=below:$z_2$] at(5.5,1.1) { };
	\node[main, fill =black!10] (bz) [label=left:$b^z$] at(2,1.1) { };
	\node[main, fill = black] (omega) [label=right:$\omega$] at(5.5,2) { };
	\node[main, fill = black] (Iota) [label=right:$\iota$] at(5.5,0) { };
	\node[main, fill = black] (Omega) [label=right:$\Omega$] at(4,2) { };
	\node[main, fill =black] (Epsilon) [label=right:$\epsilon$] at(4,0) { };
	\node[main, fill = black] (Sigma) [label=left:$\sigma$] at(2,2) { };
	\node[main, fill = black] (VarSigma) [label=left:$\Sigma$] at(2,0) { };
	\path(Sigma) edge [connect] (bz)
	(VarSigma) edge [connect] (bz)
	(bz) edge [connect] (Z)
	(Ut) edge [connect] (Zt)
	(Z) edge [connect] (Ut)
	(omega) edge [connect] (Ut)
	(Omega) edge [connect] (Z)
	(Omega) edge [connect] (Zt)
	(Epsilon) edge [connect] (Zt)
	(Iota) edge [connect] (Ut)
	(Epsilon) edge [connect] (Z)
	(Zt) edge [connect] (U_{t+1})
	(omega) edge [connect] (U_{t+1})
	(Iota) edge [connect] (U_{t+1});
	\node[draw,dashed,inner sep=3mm,label=below:{t=1},fit=(Epsilon) (Omega) (Z) (Ut)] {};
	\node[draw,dashed,inner sep=3mm,label=below:{t=2},fit=(Iota) (omega) (Zt) (U_{t+1})] {};
	\end{tikzpicture}
	\caption{Gamma Chain for User Latent Factors $u_t$ (white) Interleaved by Auxiliary Variables $z_t$ (blue). Hyperparameters are in black.}
	\label{modelImage}
\end{figure}

Therefore, we add auxiliary variables $z_{m,k,t}$ to our chains while preserving the positive correlation between $u_{m,k,t}$ and $u_{m,k,t-1}$ as first suggested by \citet{cemgilgamma}. The chain (seen in Fig. \ref{modelImage}) is constructed as follows for the $k$th component of user $m$'s latent factor:
\begin{eqnarray*}
	z_{m,k,1} &\sim& Ga(\epsilon, {b_m^z}) \\
	u_{m,k,t} | z_{m,k,t}&\sim& Ga(\iota, \omega\cdot z_{m,k,t}) \\
	z_{m,k,t+1} | u_{m,k,t} &\sim& Ga(\epsilon, \Omega\cdot u_{m,k,t})
	\label{eqn:gammachain}
\end{eqnarray*}
This construction differs from prior work in the parametrization of the gamma chains. We separate the parameter $b_m^z$ (used for the initialization of each chain when $t=1$) from the transition or chaining parameters $\Omega$ and $\omega$ in contrast to \citet{cemgilgamma}. Decoupling these parameters allows us to capture the changes in user preferences $u_{m,k}$ over time independently from the initial state values. This allows us to model both the static interests (the initial state for each chain) and the temporal corrections.

However, our contribution is not limited to the change in the gamma construction but also how we incorporate these chains in our model to ensure their stability, which was not addressed in previous works despite the obvious problem it poses to the inference task.

First note that $E[u_t] = \frac{\iota}{\omega \cdot E[z_t]}$ and $E[z_t]= \frac{\epsilon}{\Omega \cdot E[u_{t-1}]}$ which compounds until $t=1$.

In fact, the correlation is thus always positive (depends on absolute values of $\epsilon$, $\iota$, $\epsilon$ and $\Omega$) whereas the skewness depends on the ratio $\frac{\iota\cdot\omega}{\Omega\cdot\epsilon}$. If the ratio $<1$ the probability mass is shifted towards the region $u_t < u{t-1}$ and the chain exhibit a systematic positive drift (the opposite is also true for a ratio$>1$).

Thus, as t increases, $E[u_t]$ and $E[z_t]$ will increase (or decrease) monotonically ($z>0$ and $u>0$) depending on the ratio raised to the power of the time index. With a ratio signficantly larger or smaller than 1 (due to posterior updates) the chains can grow too large or too small which we experienced when first attempting to use a simple Poisson factorization. 

To ensure the numerical stability of the model during the inference routine, we leverage the update of one of the compound Poisson hyper-parameters $\Lambda_t$ (further explained in the next section). Consequently, our model rescales of the contribution of the latent factors towards the generating distributions to ensure the chains don't grow uncontrollably in time.
\section{Dynamic Compound Poisson Factorization (DCPF) Model}
\label{model}
\subsection{The DCPF Generative Model}
At each time step, we represent each user $m$ as a vector $u$ of $K$ latent preferences and each item $n$ as a vector $v$ of $K$ latent attributes. Let $z$ and $w$ be the auxiliary gamma variables for the chains of $u$ and $v$, respectively. Let the variables $b_m^z$ and $b_n^w$ parametrize the initial states of the user $m$ and item $n$ chains.

One of our contributions is the usage of the compound-Poisson generating distribution \citep{hcpf}. Instead of introducing a new parameter solely to rescale the gamma chains, independently from the generating distribution, we opted for compound Poisson as it already includes a parameter that we can leverage for that purpose in addition to providing more flexibility with the type of observation data it can handle.

A compound Poisson distribution is defined as the sum of $\eta$ i.i.d. random variables each with an element distribution $p_{\psi}(\theta,\kappa)$ \citep{hcpf}. Here, the i.i.d. variables are drawn from exponential dispersion models (EDM) and $\eta$ is a Poisson-distributed random variable of mean $\Lambda$. By virtue of the additivity property of EDM distributions, a compound Poisson variable has a distribution $p_{\psi}(\theta,\eta\cdot\kappa)$, which is also an EDM distribution with scale parameter $\eta\cdot\kappa$ and natural parameter $\theta$~\citep{hcpf}. Examples of compound Poisson include zero-truncated Poisson and gamma-Poisson.

Accordingly, with a generating compound Poisson distribution of density $p_{\psi}(\theta,\eta\cdot\kappa)$, the generative process of our DCPF model is detailed in Algorithm \ref{alg:dcpf}.

\begin{algorithm}[!ht]
	\caption{DCPF Generative Model}
	\label{alg:dcpf}
	$\sigma$,$\varSigma$,$\delta$,$\varDelta$,$\omega$,$\Omega$ were introduced earlier as the parameters of the user gamma-Markov chains. $\iota$,$\epsilon$,$\hat{\iota}$,$\hat{\epsilon}$,$\hat{\omega}$,$\hat{\Omega}$ are the item chains counterparts.
	\begin{itemize}
		\item For each user $m$ from $1$ to $M$, sample the parameter of the initial state of the chain:
		$$b_m^z \sim Ga(\delta, \delta/\varDelta)$$
		For each component $k$, sample the initial user latent factor or preference ($t=1$):
		$z_{1,k} \sim Ga(\epsilon,{b_m^z})$
		$u_{1,k} | z_{1,k} \sim Ga(\iota, \omega{z_{1,k}})$
		Then, for time $t$ from $2$ to $T$, sample the states of the chain of user preferences:
		$$z_{t,k}|u_{t-1,k} \sim Ga(\epsilon, \Omega{u_{t-1,k}})$$
		$$u_{t,k} | z_{t,k} \sim Ga(\iota, \omega{z_{t,k}})$$
		\item For each item $n$ from $1$ to $N$, sample the parameter of the initial state of the chain:
		$$b_n^w \sim Ga(\sigma, \sigma/\varSigma)$$
		For each component $k$, sample the initial item latent factor ($t=1$):
		$$w_{1,k} \sim Ga(\hat{\epsilon},{b_n^w})$$
		$$v_{1,k} | w_{1,k} \sim Ga(\hat{\iota}, \hat{\omega}{w_{1,k}})$$
		Then, for time $t$ from $2$ to $T$, sample the states of the chain of item attributes:
		$$w_{t,k}|v_{t-1,k} \sim Ga(\hat{\epsilon}, \hat{\Omega}{v_{t-1,k}})$$
		$$v_{t,k} | w_{t,k} \sim Ga(\hat{\iota}, \hat{\omega}{w_{t,k}})$$
		\item For each user $m$ and item $n$ at time $t$:
		\begin{enumerate}
			\item Sample counts
			\vspace{-0.3cm}
			$$\eta_{m,n,t} \sim Poisson(\Lambda_{m,n,t} = \sum_{k}u_{m,k,t}v_{k,n,t})$$
			\vspace{-0.7cm}
			\item Sample the compound Poisson response
			\vspace{-0.3cm}
			$$y_{m,n,t} \sim p_{\psi}(\theta, \eta_{m,n,t}\kappa)$$
		\end{enumerate}
	\end{itemize}
\end{algorithm}
The main computational problem for our model is the posterior inference of the parameters of the model given a matrix of observations indexed by user, item, and time. Once the posterior is estimated, we use the DCPF generative model to recommend items to users based on their latent factors at a given time point using the expectation of the score provided by $p_{\psi}(\theta,\eta\cdot\kappa)$.
\subsection{Stochastic Variational Inference for DCPF}
\label{svi}
Given a matrix of user ratings for items across time, we wish to compute the posterior distribution of user preferences $u_{m,k,t}$ and item attributes $v_{n,k,t}$. As for many Bayesian models, the exact posterior for Poisson factorization is computationally intractable~\citep{cemgil}. Therefore, we will use variational inference, which proposes a family of distributions over the latent variables indexed by variational parameters \citep{jordan}. Variational inference estimates the set of variational parameters that place the variational distributions of the posterior closest to the true posterior in Kullback-Liebler divergence \citep{kl}. This inference problem thus becomes an optimization problem that can be scaled to large sample sizes using stochastic optimization, leading to stochastic variational inference (SVI)~\citep{svi}.

The latent variables in our model are the user preferences $u$, item attributes $v$, their corresponding auxiliary chain variables $z$ and $w$, and the rate parameters of the initial states of each chain $b^z$ and $b^w$. Our mean-field variational distribution is thus given by:
\begin{align*}
&q(u_{m,k,t}\mid \alpha^{u}_{m,k,t},\beta^u_{m,k,t})
q(z_{m,k,t}\mid\alpha^z_{m,k,t},\beta^z_{m,k,t})\\
&q(v_{n,k,t}\mid \alpha^{v}_{n,k,t},\beta^v_{n,k,t})
q(w_{n,k,t}\mid\alpha^w_{n,k,t},\beta^w_{n,k,t})\\
&q(b_m^z\mid {\alpha}_m^{bz}, \beta^{bz}_{m})
q(b_n^w\mid \alpha_n^{bw}, \beta^{bw}_{n})
q(s_{m,n,t}\mid \boldsymbol{\varphi_{m,n,t}})
q(\eta_{m,n,t})
\end{align*}
The variational distributions for latent factors and parameters $u$, $z$, $v$, $w$, and the parameters on the initial states of the chains, $b^z$ and $b^m$, are all Gamma-distributed with independent variational shape and rate parameters $\alpha$ and $\beta$.

The variational distribution of $s$ is a multinomial distribution with parameter $\varphi$. $s$ and $\varphi$ are auxiliary variables for each user--item--time tuple typically used in PF to facilitate the model inference~\citep{cemgil}.

We optimize the variational parameters with coordinate ascent which iteratively optimizes each variational parameter while holding the others fixed at their previous approximations \citep{svi}. We assert convergence when there is $<10^{-5}$ change in predictive likelihood on a validation set.

Given DCPF's conjugacy, we derived closed-form updates for each of our parameters as shown in Algorithm~\ref{alg:svi}. Note $\tau$ and $\xi$ the learning rate parameters for the stochastic optimization~\citep{svi}.

\begin{algorithm}[!ht]
	\caption{SVI for DCPF}
	\label{alg:svi}
	\begin{algorithmic}
		\STATE {\bfseries Initialize:} Hyperparameters $\iota$,$\epsilon$,$\sigma$,$\varSigma$,$\omega$,$\Omega$,$\hat{\iota}$,$\hat{\epsilon}$,$\hat{\omega}$,$\hat{\Omega}$. $N=\#items$, $K=\#factors$
		\begin{align*}
		&{\alpha}_{m,t,k}^z = \iota + \epsilon && t_{u} = \tau\\
		&{\alpha}_m^{bz} = \sigma + K \cdot \epsilon
		\end{align*}
		\REPEAT
		\STATE Sample a user $m$ and item $n$ from the data set
		\REPEAT
		\STATE Compute local variational parameters for $m,n,t$
		\begin{align*}
		\Lambda_{m,n,t} &= \sum_{k} \frac{\alpha^{u}_{m,k,t} \alpha^{v}_{n,k,t}}{\beta^{u}_{m,k,t} \beta^{v}_{n,k,t}}\\
		q(\eta_{mn,t} = \eta) &\propto \exp\left \{-\kappa \eta \Psi(\theta) \right \}h(y_{m,n,t},\eta\kappa)\frac{\Lambda_{m,n,t}^{\eta}}{\eta!}\\
		\varphi_{m,n,k,t} &\propto \exp \left \{ \Psi(\alpha^{u}_{m,k,t}) - \log \beta^{u}_{m,k,t}+\Psi(\alpha^{v}_{n,k,t}) - \log \beta^{v}_{n,k,t} \right \}
		\end{align*}
		\STATE Update global variational parameters  [only showing equations for user chains]
		\begin{align*}
		{\alpha}_{m,k,t}^u &= (1-t_{u}^{-\xi})\alpha^{u}_{m,k,t} 
		+ \;\;\;\;\;&&t_{u}^{-\xi} \Big(\epsilon + \iota + N\cdot E[\eta_{m,n,t}]\cdot\varphi_{m,n,k,t})\Big) \\
		\beta^{u}_{m,k,t} &= (1-t_{u}^{-\xi})\beta^{u}_{m,k,t} 
		+ \;\;\;\;\;&&t_{u}^{-\xi} \Big(\omega\cdot\frac{\alpha^{z}_{m,k,t}}{\beta^{z}_{m,k,t}} + \Omega\cdot\frac{\alpha^{z}_{m,k,t+1}}{\beta^{z}_{m,k,t+1}} + N\cdot \frac{\alpha^{v}_{n,k,t}}{\beta^{v}_{n,k,t}}\Big)\\
		\beta^{z}_{m,k,t} &= (1-t_{u}^{-\xi})\beta^{z}_{m,k,t} 
		+ \;\;\;\;\;&&t_{u}^{-\xi} \Big(\omega\cdot\frac{\alpha^{u}_{m,k,t}}{\beta^{u}_{m,k,t}} + \Omega\cdot\frac{\alpha^{u}_{m,k,t-1}}{\beta^{u}_{m,k,t-1}}\Big)\\
		\beta^{bz}_{m} &= (1-t_{u}^{-\xi})\alpha^{bz}_{m} + \;\;\;\;\;&&t_{u}^{-\xi} \Big(\frac{\sigma}{\varSigma} + \sum_{k} \frac{{\alpha}_{m,t,k}^z}{\beta^{z}_{m,k,t}}\Big)\\
		\end{align*}
		\vspace{-0.9cm}
		\UNTIL{$t=end$ of training time period}
		\STATE Update learning rates
		\begin{align*}
		t_{u} &= t_{u} + 1    &&t_{i} = t_{i} + 1
		\end{align*}
		\STATE Update hyperparameters $\theta$ and $\kappa$
		\UNTIL{held-out log likelihood converges}
	\end{algorithmic}
\end{algorithm}
{\bf Computation:} The update for each user or item parameter is independent of all the other users or items, respectively. Accordingly, this inference routine is easily parallelizable. As for the runtime, the per-iteration complexity can be linear in terms of non-missing entries at each time step if we train on non-missing instead of the full matrix (difference discussed extensively in HCPF \citep{hcpf}).

{\bf Gamma chain stability:} The variational parameters for $u_t$ and $z_t$ depend on $\Lambda_t$. The update for $\Lambda_t$ simply equals the observation which can be large. As explained previously, such a change at time $t$ can cause the chain to grow unstable. We address this issue with two deliberate model choices. First, through compounding, the contribution of the variational parameter for $\Lambda_t$ towards the observations can be controlled to stay small which reflects on the updates for the variational parameters of $u$ and $v$ (see \citet{hcpf} for the interaction between $\Lambda_t$ and the latent factors).
Second, we carefully initialize the hyperparameters to control the skewness of the chain, at least for the first iteration of the posterior update.
\vspace{-0.4cm}
\section{Results}\label{results}
We consider two types of time-stamped ratings data sets: explicit ratings (e.g., 1 to 5 stars) and implicit ratings (e.g., number of song plays or clicks). We fit our model to these data sets on a fixed number of time windows (indicated in Table~\ref{table:datasets}) and then predict the ratings for the held-out time windows. We compare the prediction accuracy versus HCPF \citep{hcpf} and DPF \citep{dpf}.
\subsection{Data Sets and Discretization}
\begin{enumerate}
	\item \textbf{Active Netflix}: A subset of the Netflix Prize data set \citep{netflix} filtered by selecting users and movies who are active between the first and last time periods(10/01/98 --12/31/05) and had a minimum amount of activity across different time periods similarly to the \citet{li} and \citet{dpf} approaches.
	\item \textbf{Active Yelp}: A subset of the Yelp Academic Challenge data set that includes customers and businesses active between 2008 and 2015.
	\item \textbf{LFM Tracks/Bands}: A Last.fm dataset of binary ratings that indicate whether or not a Last.fm user played a song during the given time period \citep{lfm}. We also consider bands as items instead of tracks.
	\item\textbf{LFMSum Tracks/Bands} Indicates the number of times a user listened to a song or band for a given time period.
\end{enumerate}
The dataset dimensions can be found in Table~\ref{table:datasets}.

We split each data set into time windows using the time stamps for each rating. We kept the duration of a time window at around 3 to 9 months based on previous works and our judgment of the average pace of change for each ratings data set. Specifically, music tastes might be evolving over weeks as new songs and albums come out. Movie tastes are often cyclical over 6 months, as movies move out of theaters and onto DVDs. Interest in specific types of businesses may be static over time windows of a year or more.
\begin{table}[h!]
	\caption{{\bf Dataset Characteristics} Number of users, items, time slices or periods used for discretization, and average sparsity per time period.}
	\label{table:datasets}
	\begin{center}
		\begin{small}
			\begin{sc}
				\setlength\tabcolsep{0.05in}
				\begin{tabular}{lllll}
					{} &   \#users &  \#items &  slices & sparsity \\
					\toprule
					active-yelp 	&905	&6721	&12 &0.999348 \\
					active-netflix	&10484	&6513	&16 &0.996527 \\
					LFMTracks  		&991	&1079037 &16 &0.999448 \\
					LFMBand			&991	&107000  &16 &0.998666 \\
				\end{tabular}
			\end{sc}
		\end{small}
	\end{center}
\end{table}

We assign the ratings of the last 1 or 2 time windows in each data set to a held-out test set in order to quantify the performance. The choice of the number of windows to hold out depends on the number of ratings in those windows that guarantees a reasonable test set size. We then randomly sample and assign $5\%$ of the ratings within this test set for the validation set.
\subsection{Baselines}
We compare the performance of DCPF to the most recent dynamic and static Poisson factorization algorithms(code provided by the corresponding authors):
\begin{itemize}
	\item \textbf{HCPF} the most recent and accurate Poisson factorization model~\citep{hcpf}. To run HCPF, which is not a dynamic model, we remove the timestamps from the training, test, and validation sets and remove repeated ratings across different time periods since HCPF cannot have more than one user-item interaction for every user-item pair. In particular, we filter duplicates in our LFM data by keeping the first instance of each user-item interaction as it signals the user's initial interest in an item (keeping the final had similar results).
	Since DCPF performs compound Poisson factorization, we compare different variants of compound Poisson generating distributions across DCPF and HCPF. 
	\item \textbf{DPF} the most recent dynamic PF algorithm, which, in prior work~\citep{dpf}, was proven to outperform state-of-the-art dynamic CF algorithms, specifically, Bayesian Probabilistic Tensor Factorization \citep{tensor} and TimeSVD++ \citep{timesvd}. DPF binarizes the ratings by default (the model was only tested on binary data in prior work and the provided code did not perform well on non-binary data).
\end{itemize}
\subsection{Initialization}
We also use the same number of latent factors ($K=70$) for both HCPF and DCPF as a compromise between tractability and predictive accuracy. In fact, we experimented with a wide range $20<K<120$ and the differences were negligible on all datasets.

We initialize the parameters of the compound Poisson distribution based on the maximum likelihood estimates computed from a small sample of non-zero ratings at $t=1$ \citep{hcpf}. We also use $E[\eta_{m,n,1}]$ to initialize the rate parameter of the initial states of the chains, $b_m^z$ and $b_n^w$, by assuming that the contribution of all $k$ latent factor components for both the user and the item is equal. Therefore, $E[b_m^z] =\Delta = E[b_n^w] = \varSigma = \sqrt{E[\eta_{m,n,1}]/K}$. 

In our tests, we fixed $\omega$ and $\Omega$ to $1$ such that the drift of the chains depends on $\frac{\iota}{\epsilon}$. We sought to keep those values small and the ratio close to 1, so we initialized $\epsilon=1.01$ and $\iota=1.01$ in order to avoid numerical instability. Accordingly, only $\epsilon$, $b_m^z$, and $\iota$ control user chains, where $b_m^z$ allows each user chain to vary differently from the other users. The hyperparameters for the items were initialized in a similar fashion.

As for the SVI hyperparameters, we use 10,000 for $\tau$ the learning rate delay and 0.7 for $\xi$ the learning rate power which is similar to the HCPF initialization \citep{hcpf}.

We initialize the variance of the Gaussian distributions of DPF with the suggested default values~\citep{dpf}. In fact, we attempted to tune the initialization but found no significant difference. We also tried a few time period lengths to run on both DCPF and DPF but the intuitive choice indicated in Table~\ref{table:datasets} had the best performance for both.
\subsection{Metrics}
To understand the quality of the fit for each model, we calculate the area under the ROC curve (AUC) \citep{auc} for DPF and 4 variants each of DCPF and HCPF: zero-Truncated Poisson (ZTP), Gamma-Poisson (GA), Gaussian-Poisson (N), and Poisson-Poisson (PO). An example of the ROC Curve can be seen in Fig. 2. The AUC compares the probability that a given user-item rating in the test set exists, after binarizing all the numeric ratings to represent whether or not the item was rated at all.

We also report the test set log likelihood for HCPF and DCPF variants fit on the full matrix similarly to the comparisons in~\citet{hcpf}.
\begin{table*}[ht]
	\caption{{\bf AUC Values for HCPF, DCPF and DPF}: Bold values indicate the best performing model across static and dynamic variants. No DPF values reported for the LFMSum datasets}
	\label{table:auc}
	\begin{center}
		\begin{normalsize}
			\begin{tabular}{l||rrrr||rrrr||rr}
				\multicolumn{1}{c}{} & \multicolumn{4}{c}{HCPF} & \multicolumn{4}{c}{DCPF} & \multicolumn{1}{c}{DPF} \\ \cmidrule(l{4pt}r{4pt}){2-5} \cmidrule(l{4pt}r{4pt}){6-9} \cmidrule(l{4pt}r{4pt}){10-10}
				\multicolumn{1}{c}{Data} & \multicolumn{1}{c}{PO} & \multicolumn{1}{c}{GA}  & \multicolumn{1}{c}{N}   & \multicolumn{1}{c}{ZTP} & \multicolumn{1}{c}{PO}  & \multicolumn{1}{c}{GA}  & \multicolumn{1}{c}{N}   & \multicolumn{1}{c}{ZTP} & \multicolumn{1}{c}{} \\
				\toprule
				Active Yelp  	&	0.740	&	0.743	&	0.744	&	0.735	&	0.827	&	{\bf 0.843}	&	0.835	&	0.832	& 	0.828	&\\ 
				Active Netflix 	&	0.864	&	0.865	&	0.864	&	0.864	&	{\bf 0.871}	&	0.867	&	0.864	&	0.864	& 	0.858	&\\ 
				LFMSumTracks 	&	0.885	&	0.884	&	0.885	&	0.883	&	0.903	&	0.900	&	{\bf 0.903}	&	0.902	&  	-	&\\ 
				LFMSumBands  	&	0.913	&	0.912	&	0.912	&	0.913	&	0.913	&	{\bf 0.916}	&	0.913	&	0.902	&  	-   &\\ 
				LFMTracks 		&	0.884	&	0.885	&	0.885	&	0.885	&	0.891	&	0.891	&	0.892	&	0.889	& 	{\bf0.900}	&\\ 
				LFMBands 		&	0.878	&	0.881	&	0.879	&	0.878	&	0.902	&	0.904	&	0.901	&	0.899	& 	{\bf0.937}	&\\ 
			\end{tabular}
		\end{normalsize}
	\end{center}
\end{table*}
\begin{table*}[ht]
	\caption{{\bf Test Set Log Likelihood Comparison of HCPF and DCPF Variants}}
	\label{table:tll}
	\begin{center}
		\begin{scriptsize}
			\begin{sc}
				\setlength\tabcolsep{0.05in}
				\begin{tabular}{lrrrrrrrrrrrr}
					& LFMTracks  & LFMBands & Active Netflix & Active Yelp & LFMSumBands & LFMSumTracks \\
					\toprule
					HCPF-PO & -0.016	   & -0.073	 	& -0.238 	& -0.149 	& -0.068	& -0.019  \\  
					\rowcolor{lightgray}
					DCPF-PO & -0.004 	   & -0.009 	& -0.027 	& {\bf -0.006} 	& -0.017	& {\bf -0.004}\\
					HCPF-GA  & -0.015  		& -0.033	& -0.228 	& -0.149 	& -0.029 	& -0.018 \\
					\rowcolor{lightgray}
					DCPF-GA  & {\bf -0.002}& {\bf-0.002} 	&-0.027 & -0.006 	& {\bf -0.011} 	& -0.005\\
					HCPF-N   & -0.016 		 &  -0.039	& -0.223	 & -0.137 	& -0.038	 & -0.020  \\
					\rowcolor{lightgray}
					DCPF-N   &  -0.002 	   & 0.002 		&-0.026 	& -0.006 & -0.014 	& -0.005\\
					HCPF-ZTP & -0.016  		& -0.086 	& -0.235 	& -0.136 	& -0.068 	& -0.020 \\
					\rowcolor{lightgray}
					DCPF-ZTP & -0.004 	   & -0.008 &{\bf-0.026} & -0.006 & -0.017 	& -0.004 \\
				\end{tabular}
			\end{sc}
		\end{scriptsize}
	\end{center}
\end{table*}

DPF, in prior work, only computes ranking metrics which is typical for collaborative filtering models for implicit data~\citep{dpf}. Therefore, to compare DCPF with DPF, we generate the top $L$ items to recommend to each user based on the predicted scores. For each user we then proceed to compute the precision @$L$ (fraction of relevant items in the top-$L$ items with $L=10$ or 100). We also calculate the normalized discounted cumulative gain (NDCG) (@10 and @100)~\citep{ndcg}. 
\section{Results}
We can see from Table~\ref{table:auc} that DCPF is the best performing algorithm, in terms of AUC on 4 out of the 6 datasets with DPF only outperforming DCPF on LFMBands and LFMTracks. Since those datasets are implicit, we have to compare the fit of DCPF and DPF on those datasets based on ranking metrics.

HCPF was outperformed by DCPF on every single dataset which validates our dynamic hypothesis regarding the user and item factors. To further confirm this hypothesis we compare the test log likelihood of each DCPF and HCPF variant.
\subsection{Comparison of HCPF and DCPF}
It is clear that, overall, DCPF outperforms HCPF in terms of AUC over all datasets (example in Fig. 2). However, if we were to pairwise compare the model on a variant-basis, we can see that it is possible for some variants of HCPF to outperform their dynamic counterparts. The ZTP static model outperforms the dynamic model on \emph{LFMSumBands}. Additionally, there is no significant difference between the AUC values on \emph{Active Netflix} for both the N and ZTP variants.

\begin{figure}[h]
	\begin{center}
	\includegraphics[scale=0.6]{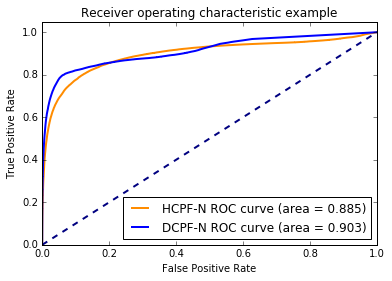}
	\label{fig:ROC}\vspace{-0.2in}
	\caption{ROC Curve Comparison on LFMSumTracks}
    \end{center}
\end{figure}
\begin{table*}[!ht]
	\caption{{\bf Ranking Metrics Comparison of DCPF and DPF} NDCG at 10 and 100, and precision at 10 and 100 are reported for DPF and DCPF}
	\label{table:dpf}
	\begin{center}
		\begin{small}
			\begin{sc}
				\begin{tabular}{llrrrr}
					Data Set & Model  &  Prec@10 &  Prec@100 & NDCG@10 & NDCG@100\\
					\toprule
					Active Yelp & DCPF  & \bf 0.007  & \bf 0.004 & \bf 0.014 & \bf 0.043 \\
					& DPF & 0.006 & 0.003 & 0.011 & 0.030 \\
					\midrule
					Active Netflix & DCPF & \bf 0.044  & \bf 0.024 & \bf 0.059 & \bf0.091 \\
					& DPF  & 0.030  & 0.017 & 0.041 & 0.077 \\
					\midrule
					LFMTracks & DCPF & \bf 0.169  & \bf 0.103 & \bf 0.193 & \bf 0.121 \\
					& DPF  & 0.056  & 0.036 & 0.063 & 0.043 \\
					\midrule
					LFMBands & DCPF & \bf 0.283  & \bf 0.165 & \bf 0.358 & \bf 0.232 \\
					& DPF  & 0.069  & 0.042 & 0.080 & 0.076 \\
				\end{tabular}
			\end{sc}
		\end{small}
	\end{center}
\end{table*}
This rare lack of improvement over the static baseline can be linked to the imbalance, in terms of number of ratings, across time windows. In fact, \emph{Active Netflix} has 10 times fewer ratings in the first window than the final one. Similarly, \emph{LFMSumBands} has 6 times fewer ratings in the first window as compared to each window in the second half of the data. In contrast, \emph{Active Yelp} is more balanced, in addition to being an active subset, and displays a much greater improvement.

This is an expected weakness of DCPF since emerging users or items (first observation at $t > 1$) can display inadequate chain updates. This is due to the great difference between the initialization and the update at $t > 1$ where a sudden shift in interest or activity cannot be perfectly captured by a smooth transition. Additionally, this kind of shift is rarely repeated over time and is the reason dynamic CF approaches use ``active" subset of the data sets~\citep{dpf, tensor, li}. However, our positive correlation might bias future updates towards higher values to match this shift even though the shift usually occurs only once when first observing a user or item.

We further compare the performances of the static and dynamic variants based on the test log likelihood. This allows us to evaluate not only the binary accuracy but also the general model fit that is more relevant to our non-binary datasets.

Accordingly, DCPF substantially outperforms HCPF for every compound-Poisson generating distribution on the \emph{Active Netflix} and \emph{Active Yelp} data sets according to Table~\ref{table:tll}. A similar margin of improvement is seen in the \emph{LFM} datasets (Table~\ref{table:tll}). This suggests that DCPF can effectively exploit data with repeated ratings over time--LastFm has repeated user-item ratings across time windows. This repeated structure allows for a more accurate estimate of the user interests and item attributes, as we have multiple opportunities to estimate parameters based on non-zero observations along the chain for a given user-item pair.
\subsection{Comparison of DCPF and DPF}
The reported AUC values for DCPF were lower than those of DPF on \emph{LFMBands} and \emph{LFMTracks}. This is to be expected on these binary datasets since DPF only samples non-missing entries and is specialized to the task of binary prediction.\citep{dpf} This advantage over sampling from the whole matrix was extensively discussed in \citet{hcpf}.

However, binary accuracy does not translate well into predicting rankings which matter more when fitting on implicit ratings datasets. In most cases, we care about modeling the scale of the interest a user has for an item. Therefore, we computed the NDCG and precision values reported in Table~\ref{table:dpf}. This should allow us to compare how well each model predicts the top 10 or 100 items for each user in the test set.

As seen in Table~\ref{table:dpf}, DCPF outperforms DPF on all datasets based on all metrics. This contrast to the AUC comparison is due to the fact that the DPF model binarizes all observations and thus ignores how strongly a user favors an item that has been rated. Additionally, long-tailed gamma priors have been proven to better capture the user and item factors in CF settings, than Gaussian priors~\citep{hpf}. This difference is clearest on the \emph{LFM} datasets which confirms the advantage of scaled ratings in assessing how much a user would rate specific items. In this case, the number of times a user plays a song is a scaled indicator of future interest in that song or similar songs.
\section{Conclusion}
In this paper, we proposed a novel conjugate and computationally stable dynamic matrix factorization (DCPF) that models the smoothly changing latent factors over time using gamma-Markov chains. Preserving the gamma-Poisson structure allows us to leverage the long-tailed gamma prior's better fit to user and item factors in sparse matrices. Our conjugate chain construction guarantees a straightforward scalable inference and potentially enables the combination with other conjugate structures or models. We also offset the main computational concern of non-negative multiplicative gamma chains, the numerical instability, with the use of Poisson compounding. This allows us to both provide flexibility for the observation data and control the gamma chains' growth.

We applied DCPF to six different time-stamped ratings data sets, and showed that DCPF achieved a higher predictive performance than a state-of-the-art static factorization model (HCPF) in terms of AUC and test log likelihood. We demonstrated an improvement over DPF in terms of ranking accuracy (NDCG and precision).

We also illustrated a drawback of this approach relative to the static model in the presence of imbalanced recommendations across time, or emerging and inactive users or items. This can be addressed with an ``active" subset of the data sets, which we already used with mitigating effects for some of the datasets. Additionally, our approach lacks the flexibility of classic time series techniques. Our smoothness assumption allows us to model the dynamic latent factors as gamma-Markov chains but ignores cyclical or seasonal behavior that can be prevalent in shopping or music listening habits.

However, we believe this conjugate, stable, scalable, and potentially modular framework will open doors for more complicated dynamic Bayesian models.

\newpage

\bibliographystyle{plainnat}
\newpage
\bibliography{nips_2016}

\end{document}